\documentclass[letterpaper]{article} 
\usepackage{aaai25}  
\usepackage{times}  
\usepackage{helvet}  
\usepackage{courier}  
\usepackage[hyphens]{url}  
\usepackage{graphicx} 
\usepackage{natbib}  
\usepackage{caption} 
\usepackage{algorithm}
\usepackage{algorithmic}

\usepackage{newfloat}
\usepackage{listings}
\DeclareCaptionStyle{ruled}{labelfont=normalfont,labelsep=colon,strut=off} 
\floatstyle{ruled}
\newfloat{listing}{tb}{lst}{}
\floatname{listing}{Listing}
\newcommand{\eg}{\emph{e.g.}}
\newcommand{\ie}{\emph{i.e.}}
\usepackage{amsmath}
\usepackage{subcaption}
\usepackage{multirow}
\usepackage{multicol}
\usepackage{pdfpages}
\usepackage{booktabs}
\usepackage{amssymb}

\usepackage{color} 

\usepackage{hyperref}
\usepackage{url}

\definecolor{custompink}{RGB}{255,105,180}
\title{FaceMe: Robust Blind Face Restoration with Personal Identification}
\author{
    Siyu Liu\textsuperscript{\rm 1}\equalcontrib \quad 
    Zheng-Peng Duan\textsuperscript{\rm 1}\equalcontrib \quad
    Jia OuYang\textsuperscript{\rm 2} \quad 
    Jiayi Fu\textsuperscript{\rm 1} \\
    Hyunhee Park\textsuperscript{\rm 3}\quad
    Zikun Liu\textsuperscript{\rm 2}\quad 
    Chun-Le Guo\textsuperscript{\rm 1, \rm 4}\thanks{Corresponding author.}\quad
    Chongyi Li\textsuperscript{\rm 1, \rm 4} \\ 
}
\affiliations{
    \textsuperscript{\rm 1}VCIP, CS, Nankai University \quad
    \textsuperscript{\rm 2}Samsung Research, China, Beijing (SRC-B)\\
    \textsuperscript{\rm 3}The Department of Camera Innovation Group, Samsung Electronics\\
    \textsuperscript{\rm 4}NKIARI, Shenzhen Futian \\ 
    
    liusiyu29@mail.nankai.edu.cn,
    adamduan0211@gmail.com, 
    jia.ouyang@samsung.com,
    fujiayi@mail.nankai.edu.cn, \\
    \{inextg.park, zikun.liu\}@samsung.com,
    \{guochunle, lichongyi\}@nankai.edu.cn\\

    \vspace{0.2cm}
    \textbf{Project page:} \textcolor{custompink}{\texttt{https://modyu-liu.github.io/FaceMe\_Homepage/}}
    
    \vspace{-0.1cm}

}

\begin{document}
\maketitle

\begin{figure*}[t]
    \centering
    \includegraphics[width=\textwidth]{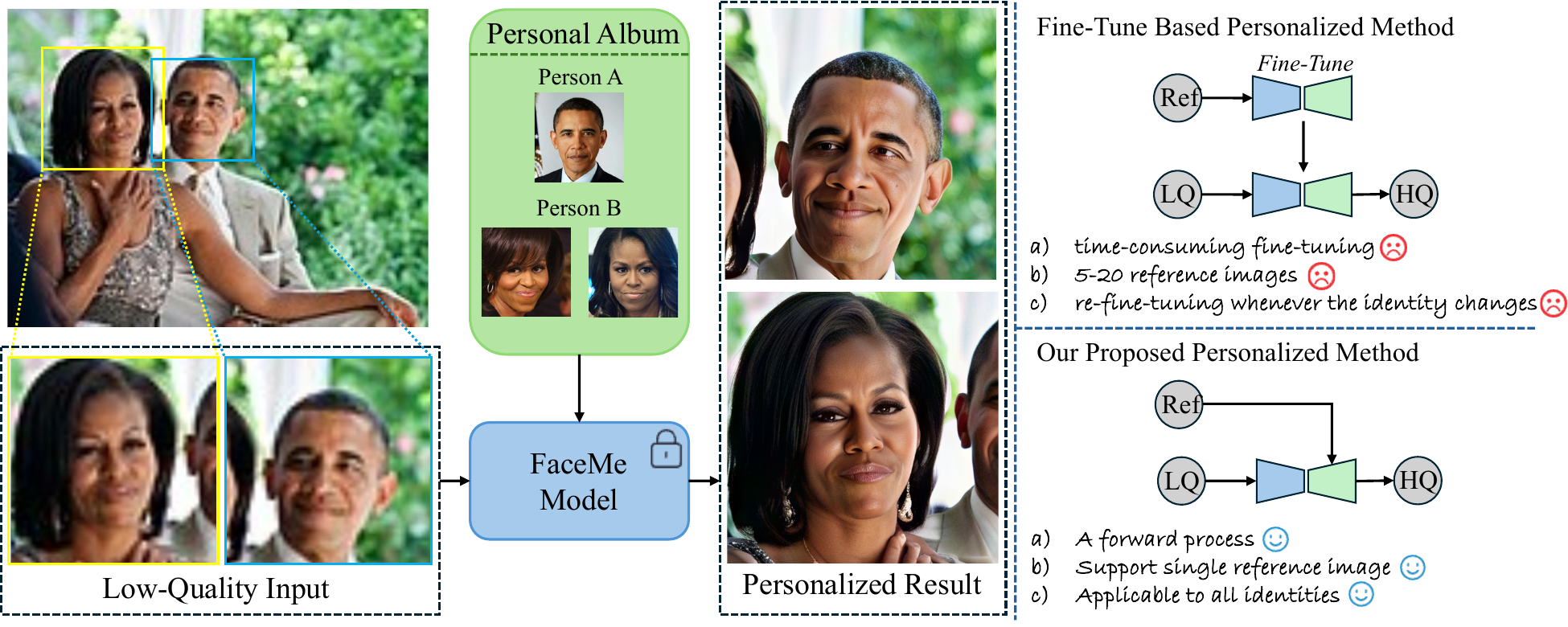}
    \vspace{-6mm}
    \caption{Make the people in the photo look like you and those you are familiar with. Using a single or a few reference images, we can restore realistic images without any fine-tuning for identity. \textbf{Zoom in for best view.}}
    \label{fig:teaser}
    \vspace{-4mm}
\end{figure*}

\begin{abstract}
\vspace{-1mm}
Blind face restoration is a highly ill-posed problem due to the lack of necessary context. Although existing methods produce high-quality outputs, they often fail to faithfully preserve the individual's identity. In this paper, we propose a personalized face restoration method, FaceMe, based on a diffusion model. Given a single or a few reference images, we use an identity encoder to extract identity-related features, which serve as prompts to guide the diffusion model in restoring high-quality and identity-consistent facial images. By simply combining identity-related features, we effectively minimize the impact of identity-irrelevant features during training and support any number of reference image inputs during inference. Additionally, thanks to the robustness of the identity encoder, synthesized images can be used as reference images during training, and identity changing during inference does not require fine-tuning the model. We also propose a pipeline for constructing a reference image training pool that simulates the poses and expressions that may appear in real-world scenarios. Experimental results demonstrate that our FaceMe can restore high-quality facial images while maintaining identity consistency, achieving excellent performance and robustness.
\vspace{-2mm}
\end{abstract}

\vspace{-4mm}

\section{Introduction}

Face restoration focuses on improving the quality of facial images by removing complex degradation and enhancing details. In the real world, facial images frequently suffer from complex degradation, \eg, blurring, noise, and compression artifacts, which can significantly impact downstream tasks like face recognition and detection.

Face restoration is inherently a highly ill-posed task because a single low-quality input can correspond to many potential high-quality counterparts scattered throughout the high-quality image space. Most advanced face restoration methods cannot guarantee identity consistency, \ie, the identity of the restored portrait may significantly deviate from the actual identity. Such deviations in identity are unacceptable in practical applications, \eg, restoring facial images in personal photo albums.

Recently, some studies have used high-quality reference images of the same identity to enhance identity consistency in restored facial images. These approaches improve results through feature alignment \cite{li2018learning,li2020enhanced,li2022learning} or by fine-tuning the diffusion model to constrain the generation prior \cite{varanka2024pfstorer,ding2024restoration,chari2023personalized}. Although these methods have achieved significant results, they have obvious limitations. The feature alignment-based methods are influenced by the degradation of the input image and the pose of the reference image. If the features are not well aligned, the quality of the restoration can significantly deteriorate. The fine-tuning-based methods typically require 5$\sim$20 reference images and need to be fine-tuned again whenever the identity changes, making it a time-consuming process. Additionally, the quality of the reference images may significantly impact image restoration quality. We illustrate the differences between our proposed personalized method and the fine-tuning-based personalized method in Fig. \ref{fig:teaser}.

In this paper, we proposed FaceMe, a fine-tuning-free personalized blind face restoration method based on the diffusion model. Given a low-quality input and either a single or a few high-quality reference images of the same identity, FaceMe restores high-quality facial images and maintains identity consistency within seconds. Remarkably, changing identities does not require fine-tuning, and the reference images can have any posture, expression, or illumination. Furthermore, the quality of the reference image does not significantly impact the quality of the restored image. To our knowledge, this is the first approach that leverages diffusion prior for personalized face restoration tasks, which does not require fine-tuning when changing identity.

The fine-tuning-free personalized blind face restoration currently faces three significant challenges: \textit{1) Influence of identity-irrelevant features from reference images; 2) Balancing the dependency between low-quality input and reference images;  and 3) Insufficient datasets for personalized blind face restoration.}

To address the \textbf{first challenge}, we propose to use an identity encoder to extract identity-related features and a multi-reference image training approach to minimize the impact of identity-irrelevant features. 
In addition, we found that the \textbf{second challenge} of balancing dependencies between low-quality input and reference images arises from simultaneous personalization and restoration training. To address it, we propose a two-stage training approach that effectively balances these dependencies. In the training stage \uppercase\expandafter{\romannumeral1}, we focus on training the model's personalization ability, \ie, training the proposed identity encoder. In the training stage \uppercase\expandafter{\romannumeral2},  we fix the identity encoder and train a ControlNet, which corresponds to the image restoration capability of the model. 
Considering the \textbf{third challenge} of insufficient training data, we propose to use synthetic datasets as reference images. Specifically, to better align reference images with real-world application scenarios, we create a pose-reference pool categorized by poses and expressions. During the synthesis process, we sample several images from this pose-reference pool to serve as pose references and select one image from a publicly available facial image dataset to serve as identity control. Using a personalized generative model, we synthesize reference images that match the poses of the given pose references while maintaining the identity of the provided identity control image, thereby producing a set of images with the same identity but diverse poses.

\section{Related Work}
\subsection{Blind Face Restoration}
Recently proposed blind face restoration methods rely on generating priors to achieve high-quality image restoration. Methods based on GAN priors, such as GPEN \cite{yang2021gan} and GFP-GAN \cite{wang2021towards}, enhance the quality of restored images by embedding rich generative priors from pre-trained face GANs into their network structures. Due to the remarkable success of VQGAN \cite{esser2021taming} in the field of image generation, researchers have been inspired to propose restoration methods utilizing vector quantization. VQFR \cite{gu2022vqfr}, CodeFormer \cite{zhou2022towards}, and RestoreFormer \cite{wang2022restoreformer} all focus on learning high-quality codebooks and achieving superior image restoration by matching high-quality vectors. More recently, the powerful generation prior of diffusion models has been employed for face restoration. DR2 \cite{wang2023dr2} leverages the diffusion model as a robust degradation removal module and employs image restoration techniques to improve the restored image's quality. DifFace \cite{yue2022difface} establishes a posterior distribution from observed low-quality images to high-quality images. By applying this distribution to each step of the denoising process of a pre-trained diffusion model, it gradually converts low-quality images into high-quality images. Although these blind face restoration methods achieve impressive results, restored identities often deviate from the real ones due to the lack of identity information.
\vspace{-2mm}
\subsection{Reference-based Face Restoration}

To enhance the quality of restored images and ensure identity similarity, several methods leverage reference images of the same identity for face restoration. GFRNet \cite{li2018learning}, ASFFNet \cite{li2020enhanced}, and DMDNet \cite{li2022learning} require alignment modules to align the features of low-quality images and reference images. However, when low-quality images' degradation is severe or the quality of the reference images is suboptimal, the restoration quality is compromised. Recently, some methods \cite{chari2023personalized,varanka2024pfstorer,ding2024restoration} achieve personalized face restoration based on diffusion models. Their core idea is to utilize $5\sim 20$ reference images of the same identity to constrain the generation priors of the diffusion model through fine-tuning. Although these methods are effective in maintaining identity, they are generally time-consuming and require re-fine-tuning whenever the identity is changed.

\begin{figure*}[!t]
	\centering
	\includegraphics[width=1\textwidth]{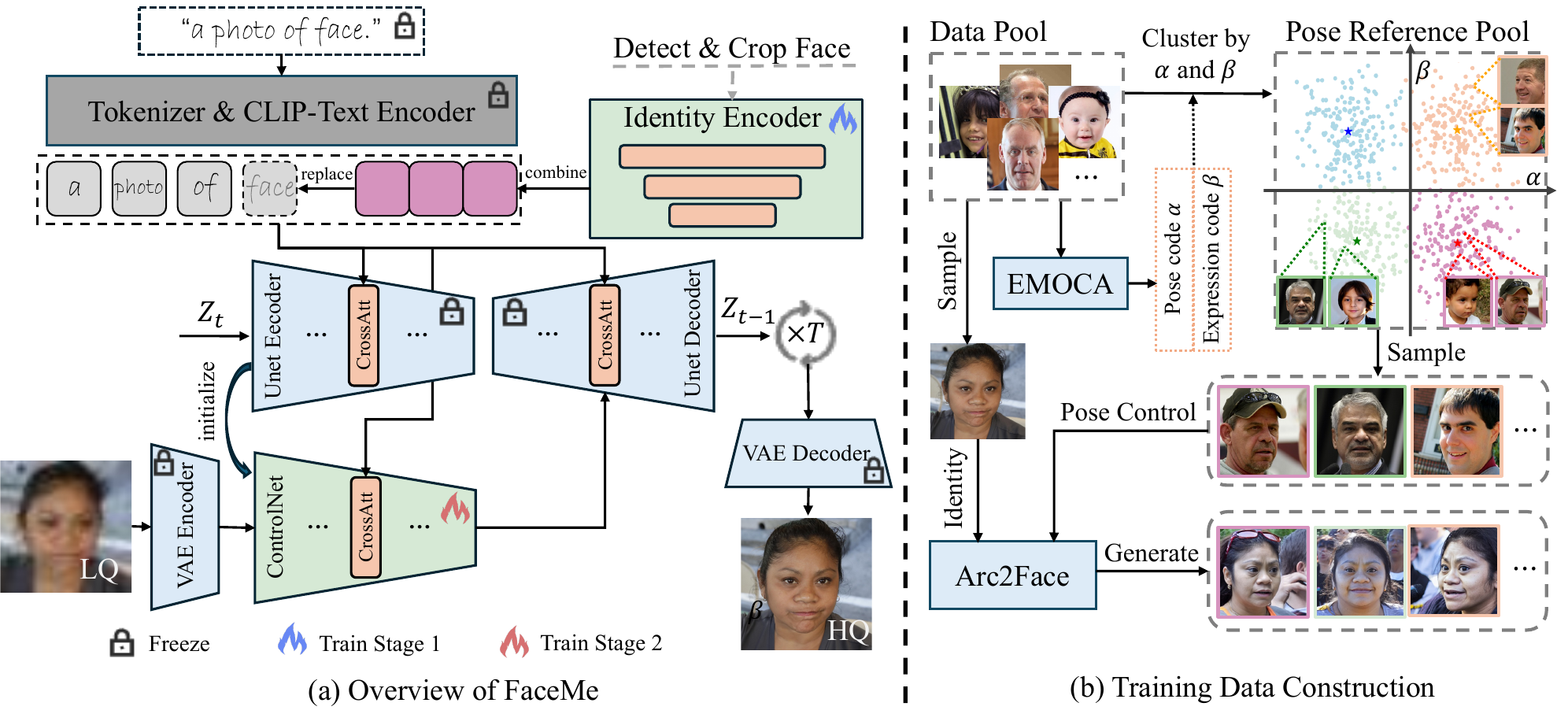}
        \vspace{-7mm}
	\caption{\textbf{Overview of proposed FaceMe (left) and training data construction pipeline (right).} For the proposed FaceMe, identity-related features from the reference image are extracted by the identity encoder, by simply combining to support multi-reference image inputs. We first use a fixed text, \ie, \textit{a photo of face.} and then apply the combined identity-related features to replace the \textit{face} embedding. The updated embeddings are sent to the cross-attention layer of the diffusion model to guide personalized face image restoration.} 

	\label{fig:overview}
	\vspace{-4mm}
\end{figure*}

\subsection{Personalized Generation}

Recently, some personalized face generation models have been proposed. 
Due to the constraints of the dataset, IP-Adapter \cite{ye2023ip}, InstantID \cite{wang2024instantid}, and Arc2Face \cite{papantoniou2024arc2face} are trained in a self-supervised manner, \ie, using the same facial image as reference and ground truth. Although this training strategy can generate realistic images that match the identity of a given reference facial image, it also introduces many identity-irrelevant features, such as posture and expressions. Photomaker \cite{li2024photomaker} collects a large-scale set of images of the same identity and supports the input of any number of reference images.

The essence of personalized generation lies in blending reference images with text descriptions to create character images that both align with the identity of the reference images and adhere to the text descriptions. Personalized image restoration, on the other hand, seeks to balance the reliance on both reference images and low-quality input, aiming to restore high-fidelity facial images that maintain consistency with the identity of the reference image. Given the similarities in the core objectives of these tasks, personalized generation methods can be effectively applied to personalized restoration.

\section{Methodology}

\subsection{Overview}

An overview of the proposed FaceMe is shown in Fig. \ref{fig:overview}. Given a few reference images of the same individual as the target image, the primary objective of this work is to leverage prior knowledge from a diffusion model to restore realistic facial images from degraded ones, while ensuring the individual’s identity in the restored images. By utilizing the identity encoder to extract identity-related features, we can obtain a unified identity representation from a given reference image without needing to fine-tune the model for each new identity. By simply combining identity-related features, our method enables an arbitrary number of reference images (single or a few). Furthermore, to effectively train our model, we design a dataset construction pipeline to remedy the lack of personalized facial image restoration datasets. 

\subsection{Problem Definition}
Let $X$, $Y$, $\mathcal{D}$, and $\mathcal{G}$ denotes the degraded facial image, the corresponding high-quality facial image, the degradation function, and the generation function, respectively. The objective of facial image restoration is to generate \(\hat{Y} = \mathcal{G}(X)\), while ensuring the following two constraints are satisfied:
\begin{equation}
    Consistency:\mathcal{D} \left( \hat{Y} \right) \equiv X,\  Realness:\hat{Y} \sim q\left( Y \right) ,\ 
\end{equation}
where $q(Y)$ denotes the distribution of high-quality facial images.

Let $Ref$ denote the reference images of the same identity as $Y$. Based on reference images, personalized face restoration can be formulated as $\hat{Y}=\mathcal{G}(X|Ref)$. In addition to meeting \textit{Consistency} and \textit{Realness}, it also needs to meet the following constraint:

\begin{equation}
    Identity\  Consistency:\  \hat{Y} \sim ID\left( Y \right),
\end{equation}
where $ID\left( Y \right)$ denotes the distribution of high-quality facial images of the same identity as $Y$.

If we consider only \textit{Realness} and \textit{Identity Consistency}, it becomes a personalized face generation task. From the perspective of problem definition, we can decompose the problem of personalized face restoration into a combination of personalized face generation and face restoration.
Next, we detail how we achieve personalized face restoration based on personalized face generation.

\subsection{Proposed Personal Face Restoration}

\subsubsection{Identity encoder}

Following recent works \cite{wu2024infinite,guo2024pulid}, we combine the CLIP \cite{radford2021learning} image encoder $\varepsilon$ and the facial recognition module $\psi$, \ie, ArcFace \cite{deng2019arcface} network, to extract identity features from facial images. Due to the different feature dimensions extracted by $\varepsilon$ and $\psi$, we employ MLPs to map the features extracted by $\psi$ to align with those extracted by $\varepsilon$. The final embeddings, which correspond to the identity information of the given reference images, are obtained by merging these two sets of features using MLPs.

Specifically, let $\{ R_{i}\}_{i=1}^{N}$ denote the given $N$ reference facial images. After face detection and cropping, $r_{i}$ is obtained from $R_{i}$. We use $\varepsilon$ and $\psi$ to extract features from $r_{i}$, denoted as $f_{i} \in \mathbb{R}^{d}$ and $g_{i} \in \mathbb{R}^{512}$, respectively, where $d$ denotes the dimension of cross-attention in the diffusion model. We then employ MLPs to align the dimension of $g_{i}$ with $f_{i}$, resulting in $\hat{g}_{i} \in \mathbb{R}^{d}$. Finally, we use MLPs to fuse $\hat{g}_{i}$ and $f_{i}$, resulting in ${s}_{i} \in \mathbb{R}^{d}$. 

\subsubsection{Combining and replacing}
 We begin by combining the extracted $\mathbf{e}_{i}$, obtaining: 
\begin{equation}
    s=Concat([s_1, \dots, s_N]), s \in \mathbb{R}^{N\times d}.
\end{equation}

Since our approach does not require semantic guidance, we consistently use a simple prompt: ``a photo of face.'', during both the training and inference phases. Let $c_{text}=\{e_1, e_2, e_3, e_4, e_5\}$ represent the embedding of the text obtained through the tokenizer and CLIP text encoder. We replace the embedding $e_4$ corresponding to the word ``face" with $s$, resulting in $c_{id}=\{e_1, e_2, e_3, s, e_5\}$, which is used as prompt embedding to guide the diffusion model for personalized face image restoration.

\subsubsection{Training strategy}

The model consists of two trainable modules, \ie, ControlNet and ID encoder (Identity Encoder in Fig. \ref{fig:overview}). 
In this work, we propose a two-stage training strategy. In training stage \uppercase\expandafter{\romannumeral1}, we simultaneously train ControlNet and ID encoder, but only save ID encoder's weights. In training stage \uppercase\expandafter{\romannumeral2}, we fix the ID encoder and only train ControlNet. In this traning process, we randomly replace identity embedding $c_{id}$ with non-identity embedding $c_{text}$ with a $50\%$ probability. The loss function for the two training stages can be expressed as:
\begin{equation}
    \mathcal{L}=\mathbb{E}_{z_{t}, I_{LQ}, I_{\text {ref}}, \epsilon}\left\|\epsilon-\epsilon_\theta\left(z_t, I_{L Q}, I_{ref}\right)\right\|_2,
\end{equation}
where $\epsilon$ is target noise, $\epsilon_\theta$ is the proposed model, $z_t$ is the latent code at time $t$,  $I_{LQ}$ is the low-quality input, and $I_{ref}$ are the reference images. Next, we will discuss the reasons for using such a training strategy.

Our training strategy is based on the following experimental insights: 1) When training the ID encoder independently, \ie, training a personalized generative model, we encounter unsatisfactory results. The ability to generate personalized faces is significantly compromised due to insufficient training data and the absence of semantics. 2)
Simultaneously training the ControlNet and ID encoder can lead to instability. As training progresses, although the model’s generative ability improves, the control ability of low-quality image diminishes, leading to undesired artifacts in restored images. However,  we found that upon completing the simultaneous training, the personalized face restoration model can seamlessly transition into a personalized generative model by simply removing the ControlNet.

The primary goal of training stage \uppercase\expandafter{\romannumeral1} is to train the ID encoder, \ie, to train the model's \textbf{personalized generation} ability. Training with ControlNet, as opposed to training the ID encoder only, can compensate for the limitations of insufficient training data and the absence of semantics. Since ControlNet provides the layout information of the target image, the ID encoder can focus on guiding the model to restore identity-specific detail features. 

In training stage \uppercase\expandafter{\romannumeral2}, we focus on balancing the dependency between low-quality input and reference images. When training the ControlNet and ID encoder simultaneously, the continuous updating of the ID encoder's weights prevents the ControlNet from providing stable low-quality image control. To enhance the ControlNet's ability, we fix the ID encoder obtained in stage I and then train only the ControlNet, replacing $c_{id}$ with $c_{text}$ with a $50\%$ probability.

\subsubsection{Inference strategy}

Following \cite{wu2024seesr}, we embed the low-quality input directly into the initial random Gaussian noise according to the training noise scheduler. Using Classifier-free guidance (CFG) \cite{ho2022classifier} for personalized guidance, the CFG in the inference stage can be represented as:
\begin{equation}
    \begin{gathered}z_{t-1}^{id}=\phi \left( z_{t},z_{LQ},c_{id} \right) ,\  z_{t-1}=\phi \left( z_{t},z_{LQ} \right) ,\ \\ \bar{z}_{t-1}=z_{t-1}+\lambda_{cfg} \times \left( z_{t-1}^{id}-z_{t-1} \right) , \end{gathered}
\end{equation}
where $\phi(\cdot)$ denotes the proposed model, $\lambda_{cfg}$ is a hyperparameter, $z_{t-1}$ is the output of model without identity control,  and $z_{t-1}^{id}$ is the output of model with identity control. In addition, to mitigate the possibility of color shift, we apply wavelet-based color correction \cite{wang2024exploiting} to the final result.

\subsection{Training Data Pool Construction}

To our knowledge, no publicly available facial dataset can support diffusion models for training with multiple reference images of the same identity. Additionally, following the Photomaker \cite{li2024photomaker} to collect a large-scale facial dataset of the same identity is both time-consuming and challenging. 
Thus, in this study, we employ synthetic facial images as reference facial images to construct our training data pool. Unlike previous methods \cite{li2018learning,li2020enhanced,li2022learning} that extract features from reference facial images and align them with degraded ones to enhance image details, our proposed method uses reference facial images as prompt to guide the diffusion model denoising process. Without a substantial shift in identity, it allows for lower-quality reference images without significantly affecting the results. Subsequently, we will detail the method of synthesizing reference facial images.

We synthesize multiple reference facial images of the same identity as the given facial image using Arc2Face \cite{papantoniou2024arc2face}, equipped with ControlNet. Given a pair of facial images $(x_{ref}, x_{pose})$, Arc2Face can synthesize facial images that maintain the identity of $x_{ref}$ and the pose of $x_{pose}$.

\subsubsection{Pose reference data pool}
We begin by constructing the pose-reference pool. Let $X=\{ x_i \}_{i=1}^{n}$ represent the images in the FFHQ dataset \cite{karras2019style}. Using EMOCA v2 \cite{danvevcek2022emoca,filntisis2022visual}, we extract the pose attribute $\theta \in \mathbb{R}^{1\times6}$and the expression attribute $\psi \in \mathbb{R}^{1\times50}$ for each image in $X$. Initially, we conduct K-Means \cite{macqueen1967some} clustering on $X$ based on $\theta$, setting the number of the cluster centers to $c_{1}$. This divides $X$ into $c_{1}$ disjointed parts, denoted as $\{X_{i}\}_{i=1}^{c_{1}}$. Subsequently, we conduct K-Means clustering on $X_i$ based on $\psi$, setting the number of the cluster centers to $c_{2}$. This results in $c_{1} \times c_{2}$ disjoint subsets of $X$, denoted as $P=\{P_{j}\}_{j=1}^{c_{1} \times c_{2}}$, forming the pose-reference pool.

\subsubsection{Same identity}
For each $x_{i}$, we randomly sample an image $p_{j}$ from $P_{j}$ to serve as the pose-reference image. Using $(x_{i}, p_{j})$ as input for Arc2Face, we synthesize the image $y_{i}^{j}$. We then assess the identity similarity between $x_{i}$ and $y_{i}^{j}$. If the identity similarity falls below $\delta$, we re-sample $p_{j}$ from $P_{j}$ for regeneration. If an acceptable $y_{i}^{j}$ is not obtained after three attempts, we stop generating it. We name the synthesized reference images for the FFHQ dataset as FFHQRef. The detailed settings can be found in the Supplementary Material.

\subsection{}\label{s3.3}

\vspace{-4mm}
\section{Experiments}
\subsection{Experimental Setup}
\subsubsection{Training datasets} 
Our training dataset consists of FFHQ dataset \cite{karras2019style} and our synthesized FFHQRef dataset, with all images resized to 512$\times$512. $I_{h}$ denotes the high-quality image from the FFHQ dataset. To form training pairs, 1$\sim$4 images with the same identity as $I_{h}$ are randomly selected from the FFHQRef dataset as reference images. The corresponding degraded image $I_{l}$ is synthesized using the following degradation model \cite{wang2021towards,zhou2022towards,yue2022difface}:
\begin{equation}
    I_{l}=\left\{ \left[ \left( I_{h}\otimes k_{\sigma} \right) \downarrow_{r} +n_{\delta} \right]_{JPEG_{q}} \right\} \uparrow_{r},
    \label{eq:degrad}
\end{equation}
where $\otimes$ denotes 2D convolution, $k_{\sigma}$ denotes a Gaussian kernel with kernel width $\sigma$, $\downarrow_{r}$ and $\uparrow_{r}$ denote downsampling and upsampling operators with scale $r$, respectively. $n_{\delta}$ denotes Gaussian noise with standard deviation $\delta$, and $[\cdot]_{{JPEG}_{q}}$ denotes the JPEG compression process with quality factor $q$. We randomly sample $\sigma$, $r$, $\delta$, and $q$ from $[0.2, 10]$, $[1, 16]$, $[0, 15]$, and $[30, 100]$, respectively.

\begin{figure*}[t]
    \centering
	\includegraphics[width=1\textwidth]{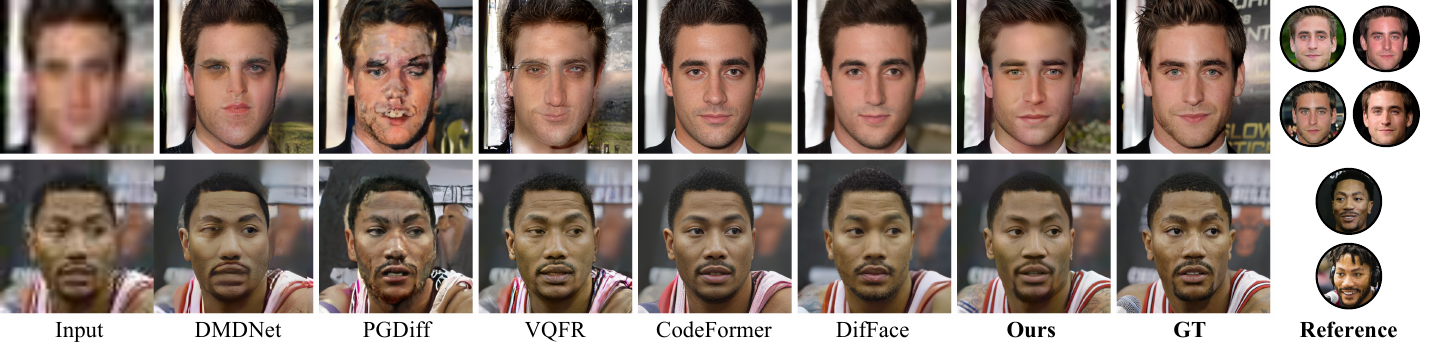}
        \vspace{-8mm}
	\caption{\textbf{Qualitative comparison on CelebRef-HQ.} In comparison to the state-of-the-art methods, our FaceMe can restore high-quality faces while maintaining identity consistency. \textbf{Zoom in for best view.} }
	\label{fig:sync_test_show1}
\end{figure*}
\begin{figure*}[t]
	\centering
	\includegraphics[width=1\textwidth]{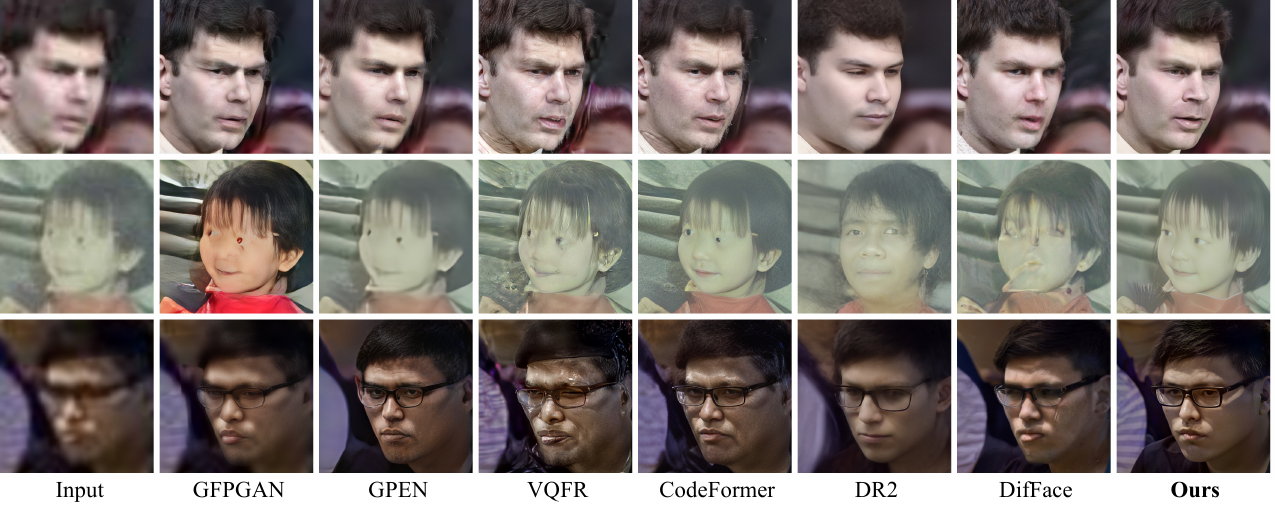}
        \vspace{-8mm}
	\caption{\textbf{Qualitative comparison on real-world faces.} The first row is from the LFW-Test; the second row is from the WebPhoto-Test; and the third row is from the Wider-Test. Our method can restore high-fidelity and high-quality images, while previous methods produce unrealistic artifacts. \textbf{Zoom in for best view.} }
	\vspace{-4mm}
	
        \label{fig:real_test_show1}
\end{figure*}
\begin{table*}[t]
    \centering
    
    \renewcommand{\arraystretch}{1.2}
    \tabcolsep=0.1cm
    \begin{tabular}{c|ccccc|cc|c|c|c|c}
    
    \toprule
        Dataset &\multicolumn{7}{c|}{CelebRef-HQ} & LFW-Test & WebPhoto-Test & WIDER-Test& \multirow{2}{*}{Ref.}\\
        Method & PSNR$\uparrow$ & SSIM$\uparrow$ & LPIPS$\downarrow$ & MUSIQ$\uparrow$ &  FID$\downarrow$ & LMD$\downarrow$ &IDS$\uparrow$ & FID$\downarrow$& FID$\downarrow$& FID$\downarrow$& \\
     
    \midrule
    
        GFP-GAN & 23.10  &  0.653 & 0.249 & 75.67 & 59.85 & 3.736 & 0.546  &51.33 & 91.51 & 40.46 & \\ 
        CodeFormer & 23.98 & 0.665 & \textbf{0.214} & \textbf{76.88} & \underline{52.21} & \underline{3.012} &  0.546 & 53.75 & 86.14 & 40.04 & \\ 
        GPEN & 24.08 & \textbf{0.686} & 0.317 & 70.01 & 72.96 & 4.197 & 0.546 &56.02 & 87.04  & 47.90 & \\ 
        VQFR & 22.75 & 0.620 & 0.262 & \underline{76.26} & 62.36 & 3.521 & 0.541 &51.83 & \textbf{77.44} & 45.02& \\ 
        DifFace & \underline{24.11} & \underline{0.679} & 0.255 & 71.85 & 53.26 & 3.293 & 0.438 & \underline{46.99} & \underline{82.65} & \underline{38.47} & \\ 
        DR2+VQFR & 22.71 & 0.650 & 0.272 & 75.66 & 64.59 & 4.017 & 0.402 & 64.44 & 117.48 & 56.84 & \\ 
        PGDiff* & 21.97 & 0.636 & 0.297 & 68.47 & 72.59 & 4.149 & \underline{0.615} & -& -& -& \checkmark\\ 
        DMDNet & 23.71 & 0.664 & 0.263 & 74.70 &  58.29 & 5.837 & 0.559 &53.34 & 89.27 & 41.98& \checkmark\\ 
        FaceMe (Ours) & \textbf{24.37} & 0.678 & \underline{0.227} & 75.62 & \textbf{51.01} & \textbf{2.908} & \textbf{0.647} & \textbf{43.95} & 92.46 &\textbf{33.59} & \checkmark \\ 
    \bottomrule

    \end{tabular}
      \vspace{-1mm}
      \caption{Quantitative comparison. The \textbf{bold} and \underline{underlined} numbers represent the best and the second best performance, respectively. * denotes that we cannot test the method on the real-world dataset due to the long inference time, which takes 90 seconds per image on an NVIDIA GeForce RTX 3090 GPU.}
      \vspace{-2mm}
        
    \label{tab:compare_data}
\end{table*}

\subsubsection{Implementation details}
We employ the SDXL model \cite{podell2023sdxl} \textit{stable-diffusion-xl-base-1.0} fine-tuned by PhotoMaker as our base diffusion model. We employ the CLIP image encoder, fine-tuned by PhotoMaker, as part of our identity encoder. We use the AdamW \cite{loshchilov2017decoupled} optimizer to optimize the network parameters with a learning rate of $5\times10^{-5}$ for two training stages. The training process is implemented using the PyTorch framework and is conducted on eight A40 GPUs, with a batch size of 4 per GPU. The two training stages are trained 130K and 210K iterations, respectively.

\subsubsection{Testing datasets}
We use one synthetic dataset CelebRef-HQ \cite{li2022learning} and three real-world datasets: LFW-Test \cite{huang2008labeled}, WebPhoto-Test \cite{wang2021towards}, and WIDER-Test \cite{zhou2022towards} for test. CelebRef-HQ is collected by crawling images of celebrities from the internet. It contains 1,005 identities and a total of 10,555 images. LFW-Test consists of 1,711 mildly degraded face images from the LFW dataset. WebPhoto-Test consists of 407 medium degraded face images from the internet. WIDER-Test consists of 970 severely degraded face images from the WIDER Face \cite{yang2016wider} dataset.  

For the synthetic dataset, we randomly select 150 identities and select one image per identity as the ground truth,  using 1$\sim$4 images of the same identity as reference images. We employ the same degradation model described in Eq.\eqref{eq:degrad} to synthesize the corresponding degraded images, maintaining the same hyperparameter settings used during training. 

For the real-world datasets, due to the lack of reference images with the same identity, we first use face restoration method, \ie, Codeformer \cite{zhou2022towards}, to restore low-quality input. The restored images are then used as input to Arc2Face to generate reference images. It is worth noting that this method of synthesizing reference images does not compromise generalization capability. The synthesized images have varying poses and relatively low quality, whereas, in real-world scenarios, the reference images provided by users are often of much higher quality.

\subsection{Comparisons with State-of-the-art Methods}

\subsubsection{Comparison methods}
We compare FaceMe with state-of-the-art methods, including GFP-GAN \cite{wang2021towards}, CodeFormer \cite{zhou2022towards}, VQFR \cite{gu2022vqfr}, GPEN \cite{yang2021gan}, DifFace \cite{yue2022difface}, and DR2 \cite{wang2023dr2}, PGDiff \cite{yang2024pgdiff}, and DMDNet \cite{li2022learning}. Since PGDiff only supports a single reference image, we use the first image in the reference dataset as the reference input during testing.

\subsubsection{Evaluation metrics}
For the synthetic dataset that contains ground truth, we adopt the following metrics for quantitative comparison: full-reference metrics PSNR, SSIM, and LPIPS \cite{zhang2018unreasonable}, as well as non-reference metrics MUSIQ \cite{ke2021musiq} and FID \cite{heusel2017gans}. We also use LMD (landmark distance using $L_{2}$ norm) and IDS$\footnote{\url{https://github.com/deepinsight/insightface}}$ (cosine similarity with ArcFace \cite{deng2019arcface}) to evaluate identity preservation. For real-world datasets without ground truth, we employ the widely-used non-reference perceptual metric FID. All evaluation metrics are measured by PyIQA$\footnote{\url{https://github.com/chaofengc/IQA-PyTorch}}$ except for LMD and IDS.

\subsubsection{Evaluation on synthetic data}
Tab. \ref{tab:compare_data} shows the performance of FaceMe on the synthetic dataset CelebRef-HQ. As shown, our method achieves the best performance in PSNR, FID, LMD, and IDS, and the second-best performance in LPIPS. Additionally, note that FaceMe has significantly improved in both LMD and IDS, which demonstrates its ability in personalization while minimizing the impact of ID-irrelevant features. The visualization results are shown in Fig. \ref{fig:sync_test_show1}. It can be observed that the compared methods either produce many artifacts in the restored images or restore high-quality images but fail to maintain identity consistency with the ground truth. In contrast, our FaceMe can restore high-quality images while preserving identity consistency.

\subsubsection{Evaluation on real-world data}
 As presented in Tab. \ref{tab:compare_data}, our FaceMe achieves the best FID score on the LFW-Test and Wider-Test datasets. LFW-Test and Wider-Test are mildly degraded and heavily degraded real-world dataset, respectively. The excellent performance on both datasets indicates that FaceMe is capable of adapting to complex degradation scenarios in the real world, demonstrating exceptional robustness. Fig. \ref{fig:real_test_show1} shows the visual comparisons of different methods. In comparison, FaceMe can handle more complex scenes and restore high-quality images without introducing unpleasant artifacts.

\begin{table}[!htb]
    \centering
    \renewcommand{\arraystretch}{1.2}
    \tabcolsep=0.1cm
    \begin{subtable}[c]{\linewidth}
        \centering
        \begin{tabular}{c|ccc|cc}
        
             \toprule
                Number & PSNR$\uparrow$ & LPIPS$\downarrow$ & FID$\downarrow$ & LMD$\downarrow$ &IDS$\uparrow$ \\ 
             \midrule   
                one & 24.65 & 0.233 & 49.38 & 2.899 & 0.620 \\ 
                two & 24.59 & 0.232 & 48.94 & 2.914 & 0.631 \\
                three & 24.54 & 0.232 & 48.52 & 2.896 & 0.632 \\
                four & 24.52 & 0.232 & 48.12 & 2.871 & 0.634  \\ 
             \midrule
               one* & 24.74 & 0.218 & 48.19 & 2.753 & 0.641 \\
               four* & 24.56 & 0.227 & 48.12 & 2.813 & 0.640 \\
               
             \bottomrule
        \end{tabular}
        \caption{Ablation study on the number of reference images. * denote using GT as part of reference.}
        \vspace{1mm}
        \label{tab:ablation1}
    
    \end{subtable}   
    
    \begin{subtable}[c]{\linewidth}
        \centering
        \begin{tabular}{c|ccc|cc}
 
             \toprule
                 Training Strategy & PSNR$\uparrow$ &  LPIPS$\downarrow$ & FID$\downarrow$ & LMD$\downarrow$ &IDS$\uparrow$ \\ 
             \midrule
               one-stage training& 22.23 & 0.274 & 61.84 & 4.195 & 0.534 \\ 
               two-stage training& 24.37 & 0.227 & 51.01 & 2.908 & 0.647  \\
             
             \bottomrule
        
        \end{tabular}
        
        \caption{Ablation study on the proposed training strategy.}
        \vspace{1mm}
        
        \label{tab:ablation2}
    \end{subtable}   
    \begin{subtable}[c]{\linewidth}
        \centering
        \begin{tabular}{c|ccc|cc}

             \toprule
                Identity Encoder & PSNR$\uparrow$ & LPIPS$\downarrow$ & FID$\downarrow$ & LMD$\downarrow$ &IDS$\uparrow$ \\ 
             \midrule
               w/o ID encoder& 24.88 & 0.281 & 59.87 & 3.083 & 0.568 \\ 
               w/ ID encoder& 24.37 &  0.227 & 51.01 & 2.908 & 0.647  \\
             \bottomrule
        
        \end{tabular}
        
        \caption{Ablation study on the proposed identity encoder.}
        
        \label{tab:ablation3}
    \end{subtable}
    \vspace{-2mm}
    \caption{Ablation studies. }
    \vspace{-4mm}
    \label{tab:ablation}
\end{table}

\subsection{Ablation Studies}
Due to space limitations, more ablation studies and the complete table for the ablation studies are provided in the Supplementary Material.

\subsubsection{The number of reference images}
We study the effects of different numbers of reference image inputs on the restored results. As shown in Tab. \ref{tab:ablation1}, as the number of reference images increases, the quality of the results improves slightly. The improvement in LMD and IDS metrics indicates that adding reference images enhances identity consistency. We attribute this improvement to the influence of the most effective reference image. When the ground truth (GT) image is used as a reference (0Ref. w GT), the model achieves its best performance. Even when the GT image is mixed with other reference images (3Ref. w GT), the model still benefits positively from GT's inclusion.

\subsubsection{Training strategy}
\begin{figure}[htb]
    \centering
    \includegraphics[width=\linewidth]{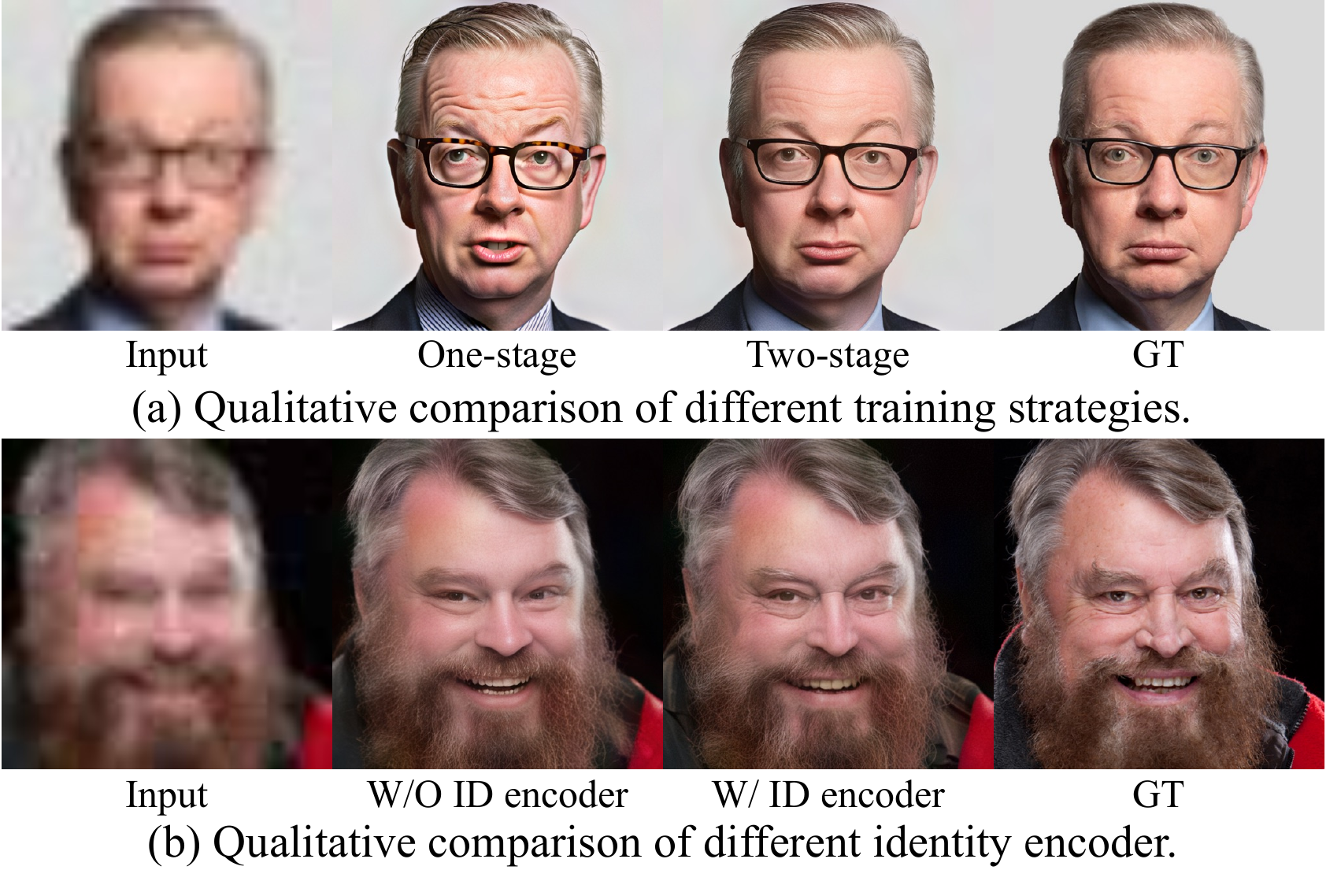}
    \vspace{-6mm}
    \caption{Qualitative comparison of ablation studies.} 
    \label{fig:Qualitative_comparison}
    \vspace{-4mm}
\end{figure}

Tab. \ref{tab:ablation2} presents the comparison of our proposed two-stage training strategy with the one-stage training strategy, \ie, jointly training ControlNet and ID encoder. While the one-stage training strategy can restore high-quality images, it significantly reduces the control over low-quality inputs and compromises identity protection. In Fig. \ref{fig:Qualitative_comparison}a, we observe that the one-stage training produces some unnatural characteristics, such as excessive wrinkles and clearly unrealistic expressions, indicating that the generative capacity is overly strong. In contrast, the two-stage training effectively balances the dependency between low-quality input and reference images.

\subsubsection{Identity encoder}
We compare the proposed identity encoder with not using an identity encoder, as shown in Tab. \ref{tab:ablation3}. Using an identity encoder can significantly improve identity consistency compared to the model without an identity encoder. Additionally, as shown in Fig. \ref{fig:Qualitative_comparison}b, using the proposed identity encoder produces clearer and higher-quality images. In contrast, the absence of an identity encoder results in blurry restored images.

\section{Conclusion}
We propose a method to address the issue of identity shift in blind facial image restoration. Based on the diffusion model, we use identity-related features extracted by the identity encoder to guide the diffusion model in recovering face images with consistent identities. Our method supports any number of reference images by simply combining identity-related features. In addition, the strong robustness of the identity encoder allows us to use synthetic images as reference images for training. Moreover, our method does not require fine-tuning the model when changing identities. The experimental results demonstrate the superiority and effectiveness of our method.

\newpage
\section{Acknowledgments}
This work is funded by the National Natural Science Foundation of China (62306153) and the Fundamental Research Funds for the Central Universities (Nankai University, 070-63243143). The computational devices of this work is supported by the Supercomputing Center of Nankai University (NKSC).

\bibliography{references}
\clearpage


\section*{Supplementary material}

# FaceMe: Robust Blind Face Restoration with Personal Identification
## — *Supplementary Material* —

In this supplementary material, we provide the following content:
- Detailed structure of the proposed identity encoder in Section 1.
- Detailed configurations for constructing the training data pool in Section 2.
- Complete ablation studies in Section 3.
- Discussion on ID-irrelevant features in Section 4.
- Discussion on the need for constructing the FFHQRef dataset in Section 5.
- Visualization of the FFHQRef dataset in Section 6.
- More real-world visual comparisons in Section 7.
- Discussion on the limitation of our proposed method in Section 8.

## 1. Identity encoder

The detailed architecture of the identity encoder is illustrated in Fig. 1. During the training phase, the weights of the CLIP encoder and ArcFace are kept frozen, while only the MLPs is trained.

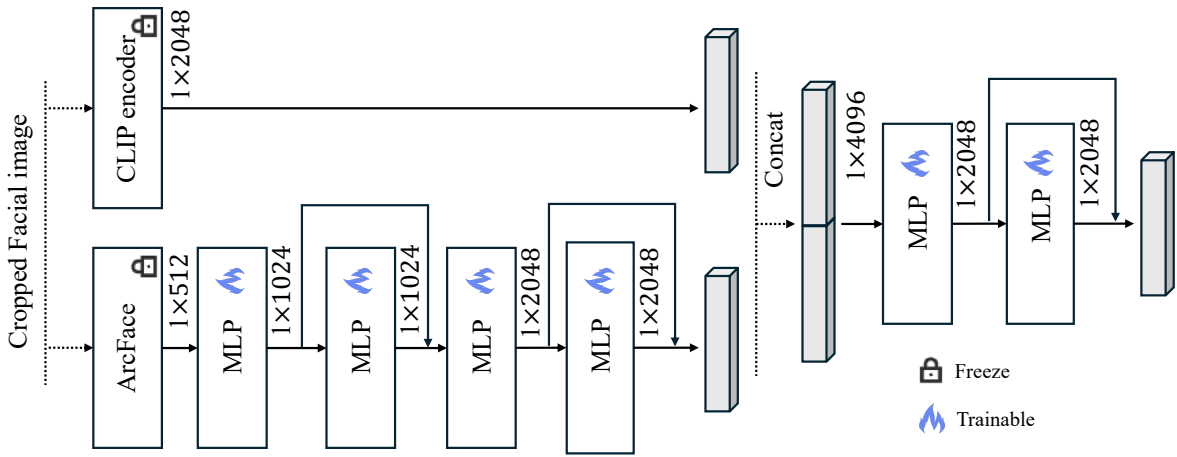

Figure 1. Architecture of the identity encoder.

## 2. Detailed configurations for constructing the training data pool.

We respectively set the number of cluster centers $c_1$ and $c_2$ to 3, resulting in a total of 9 disjoint pose reference subsets. Following the synthesis pipeline described in the main paper and applying a sequence of filtering steps, we obtained a total of 615,480 synthetic reference images. Additionally, to further enrich the reference images and address cases where a ground truth (GT) may lack a corresponding synthesized reference, we integrated the FFHQ dataset into the FFHQRef dataset, ensuring that each GT has at least one reference image.

## 3. Complete ablation studies

### 3.1. Complete table

Due to space constraints in the main paper, we are unable to present the complete experimental results. The complete table is provided in Tab. 1, corresponding to Tab. 2 in the main paper.

| Number | PSNR↑ | SSIM↑ | LPIPS↓ | MUSIQ↑ | FID↓ | LMD↓ | IDS↑ |
|---|---|---|---|---|---|---|---|
| one | 24.65 | 0.670 | 0.233 | 74.55 | 49.38 | 2.899 | 0.620 |
| two | 24.59 | 0.668 | 0.232 | 74.69 | 48.94 | 2.914 | 0.631 |
| three | 24.54 | 0.666 | 0.232 | 74.89 | 48.52 | 2.896 | 0.632 |
| four | 24.52 | 0.665 | 0.232 | 74.88 | 48.12 | 2.871 | 0.634 |
| one* | 24.74 | 0.672 | 0.218 | 74.66 | 48.19 | 2.753 | 0.641 |
| four* | 24.56 | 0.666 | 0.227 | 75.05 | 48.12 | 2.813 | 0.640 |

(a) Ablation study on the number of reference images. * donate using GT as part of the reference.

| Training Strategy | PSNR↑ | SSIM↑ | LPIPS↓ | MUSIQ↑ | FID↓ | LMD↓ | IDS↑ |
|---|---|---|---|---|---|---|---|
| one-stage training | 22.23 | 0.629 | 0.274 | 77.24 | 61.84 | 4.195 | 0.534 |
| two-stage training | 24.37 | 0.678 | 0.227 | 75.62 | 51.01 | 2.908 | 0.647 |

(b) Ablation study on the proposed training strategy.

| Dataset | CelebRef-HQ | | | | | | | LFW-Test | WebPhoto-Test | WIDER-Test |
|---|---|---|---|---|---|---|---|---|---|---|
| Identity Encoder | PSNR↑ | SSIM↑ | LPIPS↓ | MUSIQ↑ | FID↓ | LMD↓ | IDS↑ | FID↓ | FID↓ | FID↓ |
| w/o ID encoder | 24.88 | 0.706 | 0.281 | 68.55 | 59.87 | 3.083 | 0.568 | 47.44 | 98.92 | 36.64 |
| w/ ID encoder | 24.37 | 0.678 | 0.227 | 75.62 | 51.01 | 2.908 | 0.647 | 43.95 | 92.46 | 33.59 |

(c) Ablation study on the proposed identity encoder.

Table 1. Complete results of ablation studies.

### 3.2. Results under increasing levels of degradation.

As shown in Fig. 2, under the setting of using only a single reference image, our method maintains identity consistency effectively even as the level of degradation increases. While non-personalized methods can restore high-quality images in cases of severe degradation, the restored faces often deviate from their true identities. When the reference image does not match the real identity, our method relies more on the low-quality input in cases of less severe degradation. However, as the degradation level increases, the model increasingly depends on the reference image. This shows that our method effectively balances the dependency between the low-quality input and the reference image.

### 3.3. Ablation studies between synthetic reference image and real reference image.

We include an additional ablation study to investigate the impact of using synthetic reference images and real reference images on the results. Let $ref$ represent the real reference image, and we use Arc2Face [3] to generate a synthetic reference image that matches the identity and pose of $ref$. As shown in Tab. 2, it is evident that our proposed method can achieve excellent results whether using synthetic reference images or real reference images.

| Reference images type | PSNR↑ | SSIM↑ | LPIPS↓ | MUSIQ↑ | FID↓ | LMD↓ | IDS↑ |
|---|---|---|---|---|---|---|---|
| synthetic | 24.56 | 0.688 | 0.235 | 73.503 | 54.344 | 2.834 | 0.652 |
| real | 24.37 | 0.678 | 0.227 | 75.62 | 51.01 | 2.908 | 0.647 |

Table 2. Ablation study on the type of reference images.

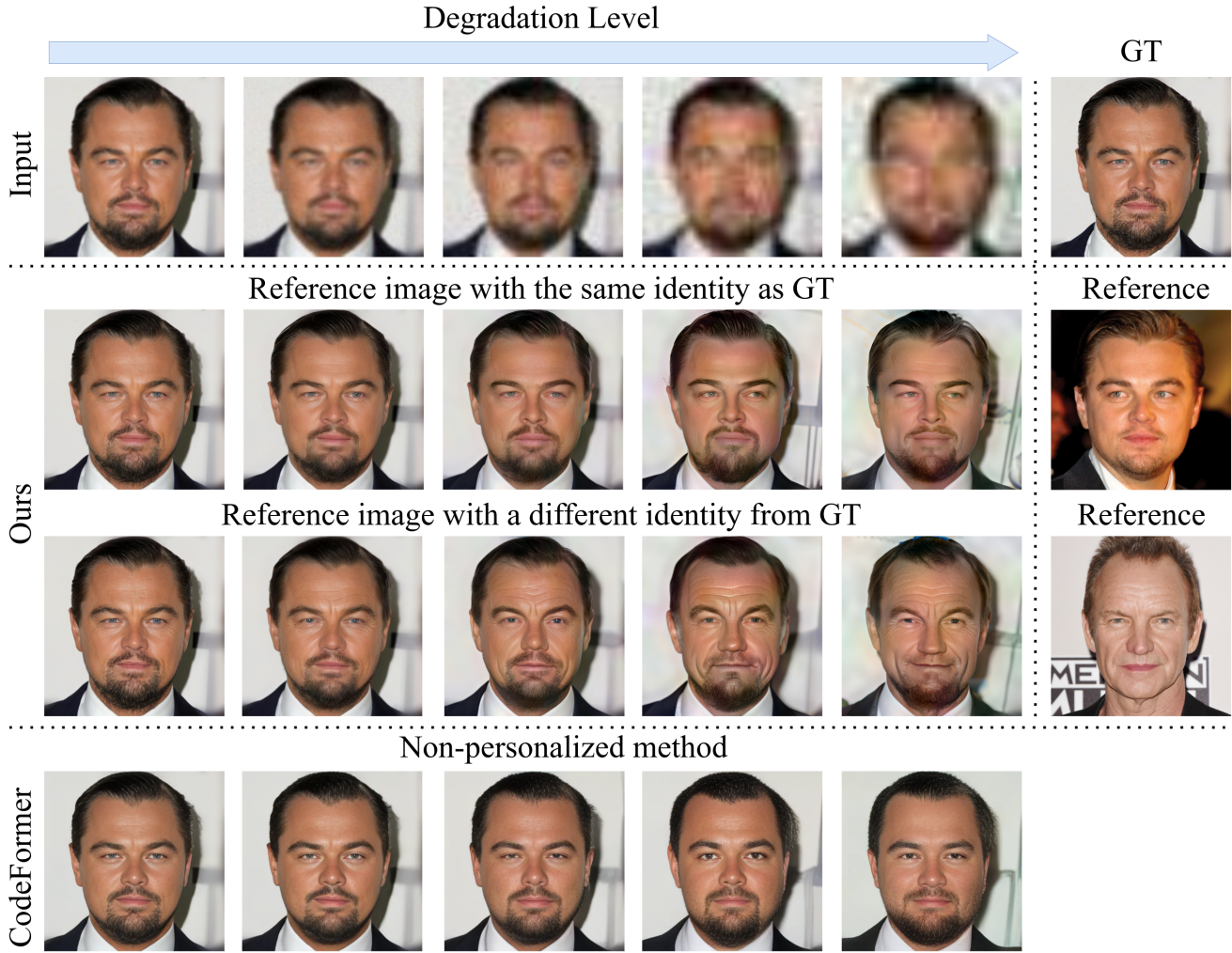

Figure 2. Qualitative comparison under increasing levels of degradation. Our method can effectively maintain identity consistency as degradation level increases, while also balancing the dependency between the low-quality input and the reference image. **Zoom in for best view.**

## 4. Discussion on ID-irrelevant features

ID-irrelevant features have less impact on identity preservation but significantly affect the results' quality. As shown in Tab. 1 of the main paper, DMDNet and PGDiff, which do not consider the impact of ID-irrelevant features, show slight improvements in identity preservation but notable drops in LMD. For example, a smile in the reference image (an ID-irrelevant feature) may transfer to the restored image. By stacking ID embeddings during training, our method minimizes the impact of ID-irrelevant features, achieving the best LMD.

## 5. The need of FFHQRef dataset

For personalized generation, studies [2–4] have shown that training a robust model requires a large dataset with diverse identities. For image restoration, high-quality GT images are crucial for optimal results. As outlined in the problem definition of the main paper, personalized face restoration can be decomposed into personalized generation and image restoration. Therefore, we need a large-scale training dataset with many identities and the corresponding high-quality GT images to support training a robust personalized face restoration model.

In previous reference-based facial restoration studies [1], the CelebHQ-Ref[1] dataset was commonly used for training. However, we have identified several limitations in CelebHQ-Ref that make it less suitable for our specific task:

1) Insufficient data. CelebHQ-Ref (10k images) is considerably smaller than commonly used face restoration datasets like FFHQ (70k) and personalized generation datasets like Laion-Face (50M) and WebFace42M (40M).

2) Inconsistent identity-related features. The CelebHQ-Ref dataset comprises photos of celebrities taken in different occasions and over varying time periods. As a result, some portrait collections span significant time intervals, leading to substantial variations in appearance even for the same individual.

3) Low quality. The image quality is not good enough to support image restoration. For example, many images have large areas of blurring caused by padding. These limitations highlight the need for a more consistent and higher-quality dataset, which leads us to generate synthetic data.

In contrast, combining the FFHQ and FFHQRef datasets can address these issues effectively. The reasons are presented as follows.

1) Abundant Data. The FFHQ + FFHQRef includes 70K images as ground truth (GT) and over 610K images as reference images, providing abundant data for training.

2) Consistent Identity Features. Reference images in FFHQRef are highly similar to their corresponding GT images in FFHQ, ensuring consistency in identity-related features.

3) High Quality. Using the FFHQ dataset as GT guarantees the quality of the training data.

In summary, the introduction of the FFHQRef dataset is highly necessary to address existing challenges and enhance the effectiveness of training.

## 6. FFHQRef dataset visualization

We present some visualization results of the FFHQRef dataset, as shown in Fig 3. Each group consists of two rows: the first image in the first row represents the GT from FFHQ, while the remaining images are synthesized using Arc2Face to construct FFHQRef.

## 7. More qualitative results

We provide more qualitative comparisons on real-world datasets, as shown in Figs. 4, 5, and 6. The LFW-Test dataset contains less severe degradations, allowing most methods to restore faces with decent quality. However, some methods struggle with maintaining fidelity, while others lack clarity. Our approach effectively balances both fidelity and clarity. The WebPhoto-Test and Wider-Test datasets contain severe degradation. Our method produces results with fewer artifacts and improved visual quality for such heavily degraded data.

## 8. Limitation

Our method supports the input of any number of reference images, but it faces a controllability issue where we cannot control the influence of a specific reference image on the results. Additionally, since our method is based on the diffusion model, it inherits its limitations, such as slow sampling speed and occasional artifacts or visual distortions. We have observed that some recent works [5] have achieved single-step diffusion, and in future work, we plan to explore training a more efficient model. Lastly, due to limitations in training data, our method relies heavily on large amounts of synthetic data. There is an inherent gap between synthetic and real data, which can introduce some bias in certain cases. Moving forward, we intend to collect a small amount of high-quality real data to fine-tune the model for improved results.

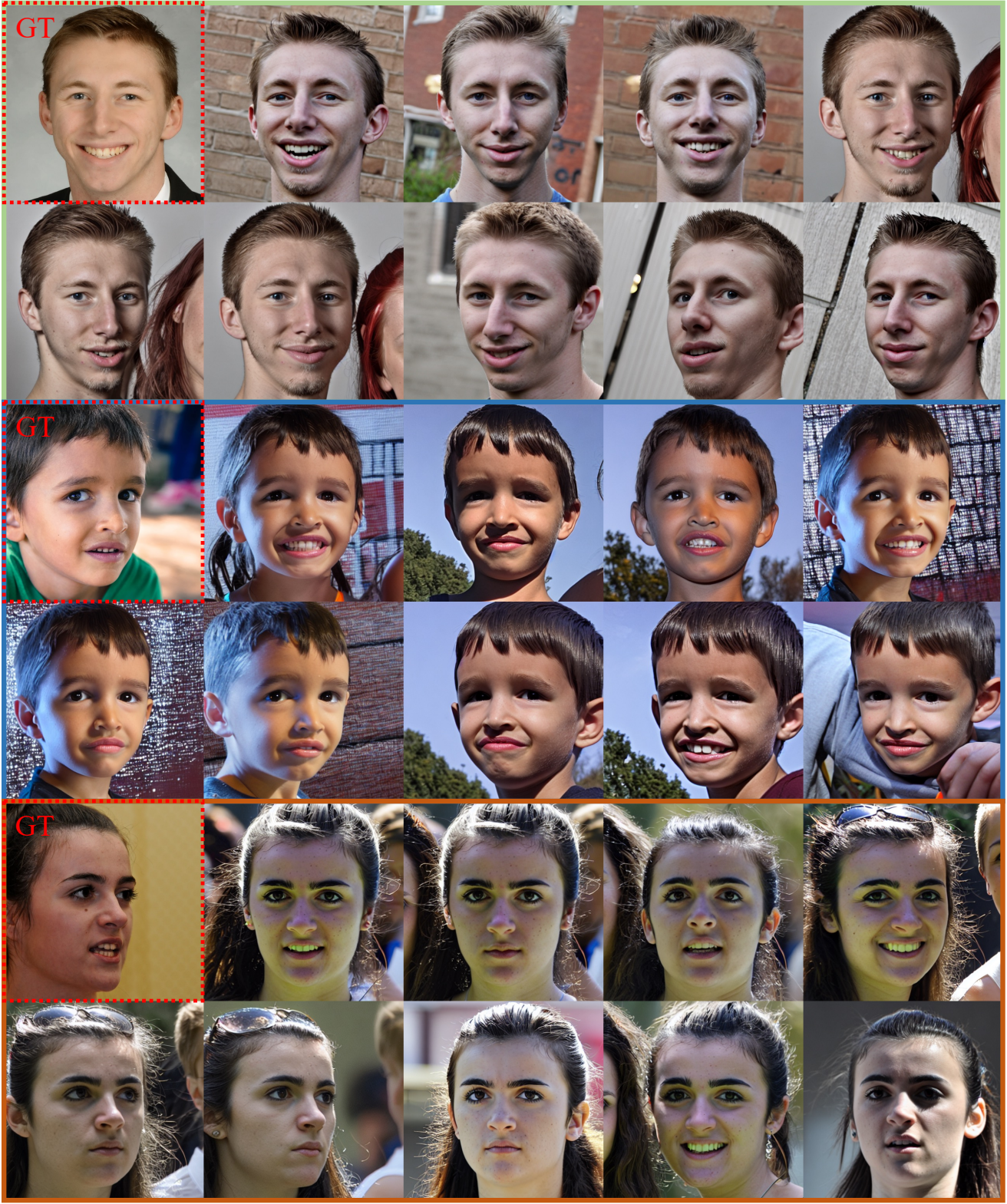

Figure 3. FFHQRef dataset visualization. **Zoom in for best view.**

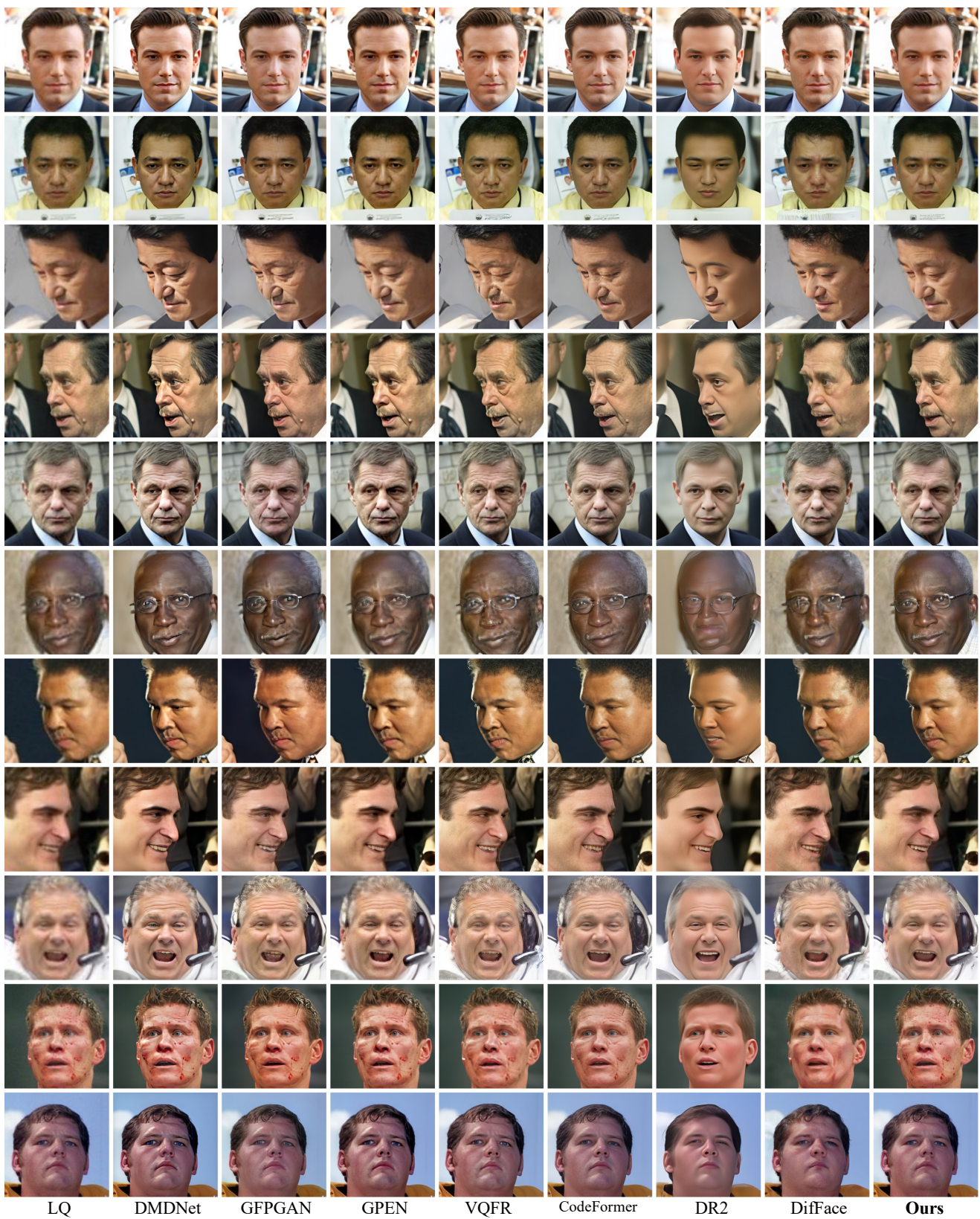

Figure 4. Qualitative comparison on LFW-Test. **Zoom in for best view.**

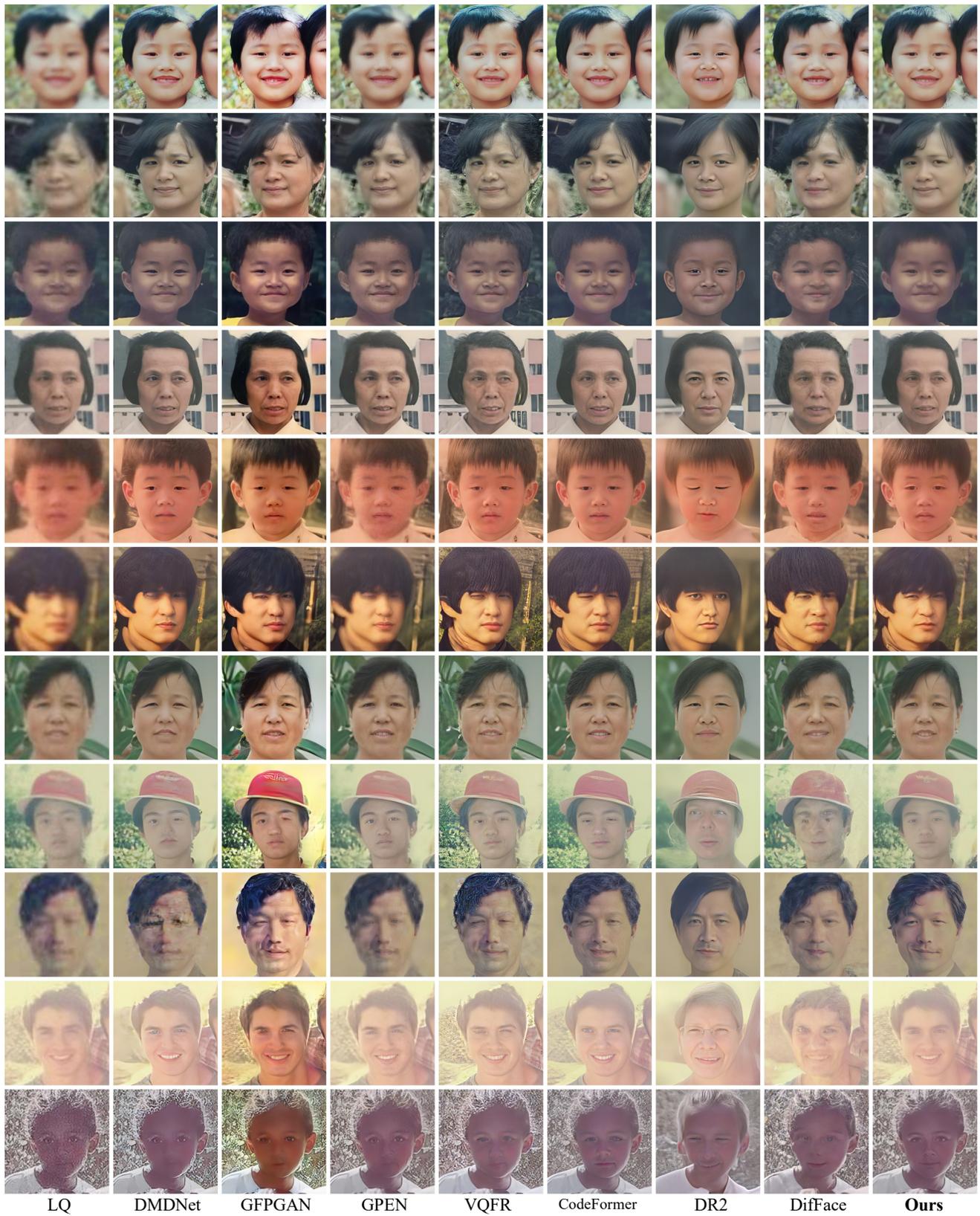

Figure 5. Qualitative comparison on WebPhoto-Test. **Zoom in for best view.**

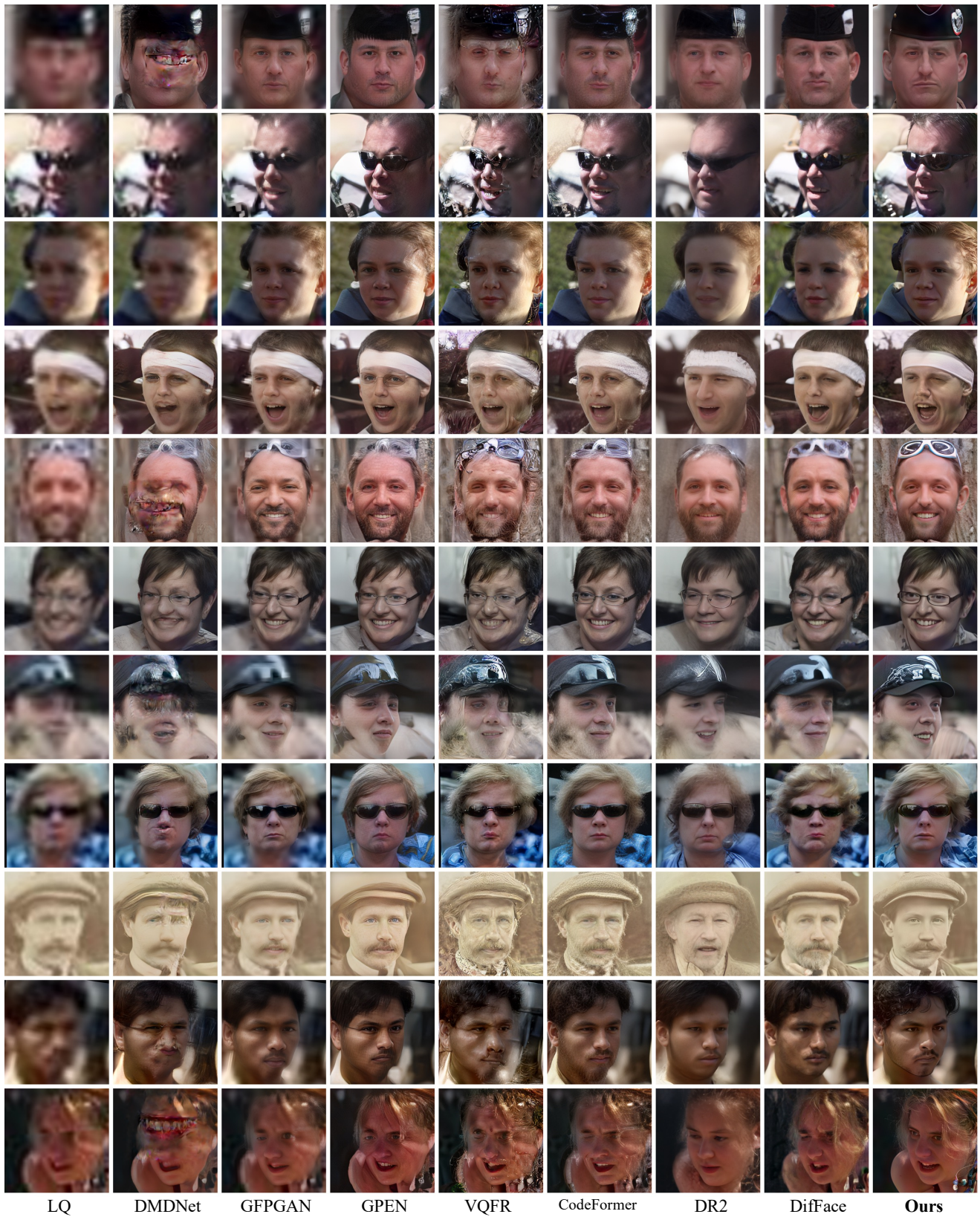

Figure 6. Qualitative comparison on Wider-Test. **Zoom in for best view.**

 
\clearpage

\end{document}